\documentclass[twocolumn,
]{ceurart}

\usepackage{lmodern}
\usepackage{listings}  

\usepackage{algorithm} 
\usepackage{algpseudocode} 
\floatname{algorithm}{Algoritmo}

\usepackage{caption}
\usepackage{url}

\usepackage{breakurl}
\renewcommand{\abstractname}{}    

\usepackage{titlesec}

\titleformat{\section}
  {\normalfont\Large\bfseries}{\thesection .}{0.2em}{}

\titleformat{\subsection}
  {\normalfont\large\bfseries}{\thesubsection .}{0.2em}{}

\titleformat{\subsubsection}
  {\normalfont\normalsize\bfseries}{\thesubsubsection .}{0.2em}{}
\begin{document}

\copyrightyear{2024}
\copyrightclause{Copyright for this paper by its authors.
  Use permitted under Creative Commons License Attribution 4.0
  International (CC BY 4.0).}

\conference{SEPLN-2024: 40\textsuperscript{th} Conference of the Spanish Society for Natural Language Processing. Valladolid, Spain. 24-27 September 2024.}

\title{%
Detección Automática de Patologías en Notas Clínicas en Español Combinando Modelos de Lenguaje y Ontologías Médicos \newline \newline 
Automatic Pathology Detection in Spanish Clinical Notes Combining Language Models and Medical Ontologies} 

\author[1]{León-Paul Schaub-Torre}[%
orcid=0000-0002-0116-9698,
email=leon.schaub@fundacionctic.org,
]
\cormark[1]
\fnmark[1]
\address[1]{CTIC Technology Centre. W3C Spain Office host, Ada Byron 39, Gijón, 33203, Asturias, Spain}

\author[1]{Pelayo Quirós}[%
orcid=0000-0002-0500-9034,
email=pelayo.quiros@fundacionctic.org,
]
\fnmark[1]
\author[1]{Helena García-Mieres}[%
orcid=0000-0002-2813-1737,
email=helenagmieres@gmail.com,
]
\cortext[1]{Corresponding author.}
\fntext[1]{These authors contributed equally.}


\begin{abstract}\\

\noindent \textbf{\normalsize Resumen} \\
En este artículo presentamos un método híbrido para la detección automática de patologías dermatológicas en informes médicos. Usamos un modelo de lenguaje amplio en español combinado con ontologías médicas para predecir, dado un informe médico de primera cita o de seguimiento, la patología del paciente. Los resultados muestran que el tipo, la gravedad y el sitio en el cuerpo de una patología dermatológica, así como en qué orden tiene un modelo que aprender esas tres características, aumentan su precisión. El artículo presenta la demostración de resultados comparables al estado del arte de clasificación de textos médicos con una precisión de 0.84, micro y macro F1-score de 0.82 y 0.75, y deja a disposición de la comunidad tanto el método como el conjunto de datos utilizado. \\\\
\noindent \textbf{\normalsize Abstract} \\In this paper we present a hybrid method for the automatic detection of dermatological pathologies in medical reports. We use a large language model combined with medical ontologies to predict, given a first appointment or follow-up medical report, the pathology a person may suffer from. The results show that teaching the model to learn the type, severity and location on the body of a dermatological pathology as well as in which order it has to learn these three features significantly increases its accuracy. The article presents the demonstration of state-of-the-art results for classification of medical texts with a precision of 0.84, micro and macro F1-score of 0.82 and 0.75, and makes both the method and the dataset used available to the community. \\

\end{abstract}


\maketitle


\section{Introducción} \label{intro}

La digitalización de informes médicos (EHR por \textit{electronic health records}) es una iniciativa internacional que lleva décadas en desarrollo. Uno de los primeros protocolos de digitalización de los informes es el ISO TC 215\footnote{\url{https://www.iso.org/committee/54960.html}} creado en 1998. Se trata de una norma cuyo objetivo es estandarizar la digitalización de los informes de más de 50 países, incluyendo España, tanto de tipo fotográfico (radiología, ecografía, etc.), como textual. Esto permite tener un contexto y un historial de cada paciente, así como facilitar los seguimientos \cite{upm231}. Sin embargo, la aceleración de esta digitalización en los últimos 15 años, con la globalización de Internet a alta velocidad y de la capacidad de los servidores, ha provocado un crecimiento de la cantidad de datos. Es por eso por lo que el procesamiento del lenguaje natural (PLN) tiene gran potencial como herramienta de ayuda a los médicos para facilitarles el trabajo de seguimiento de pacientes, al preanalizar los EHR \cite{PLN6493} extraer entidades (NER) \cite{PLN2721}, o predecir las patologías que padece una persona \cite{PLN5286}. En paralelo, los progresos en aprendizaje profundo \cite{9075398} desde los años 2010 y los transformadores a partir de 2018 \cite{NIPS2017_3f5ee243} han permitido la creación de modelos más precisos \cite{schaub2023inteligencia}. Combinando ambos avances, en los últimos años se han desarrollado modelos pre-entrenados de lenguaje especializados en el vocabulario médico, y ajustados (\textit{fine-tuned}) para las aplicaciones mencionadas \cite{DeFreitas2021,9903583,9964038}. En lengua española y en cualquier otro idioma distinto del inglés \cite{Neveol2018} los recursos existentes son limitados, pero modelos como los desarrollados por \cite{PLN6403,aracena-etal-2023-pre} consiguen resultados comparables a modelos en lengua inglesa para tareas de extracción de información, \cite{rojas-etal-2022-clinical} como es el caso de para NER.

Aún así, son pocos los trabajos que se enfocan en la predicción de una enfermedad dentro de un informe clínico \cite{app122211709}. Existen trabajos de encaje léxico que han tenido éxito para conectar un informe y un concepto (por ejemplo, enfermedad o tipo de enfermedad) \cite{Araki2023,DBLPjournals/corr/abs-2107-03134}, logrando superar a los trabajos de ontología y de semántica de la última década \cite{articleHuangetla,BUCHAN201723}. Pese a ello, apenas existen corpus de referencia para tener a la vez informes médicos en español y la patología asociada, habiendo identificado como única referencia el corpus CARES \cite{CHIZHIKOVA2023106581}, si bien está centrado en datos radiológicos. Tampoco se ha detectado un método del estado de la técnica que sea capaz de predecir a qué patología(s) corresponde un determinado informe médico.

Por otra parte, la motivación de este trabajo viene dada de no ser NER una tarea adaptada a nuestro problema por dos motivos: 
\begin{enumerate}
 \item No tenemos un conjunto de datos etiquetado en entidades nombradas (EN), lo cual supondría realizar una campaña de etiquetado y contar con un conocimiento experto del cual no disponemos.
 \item Aunque tuviéramos ese conjunto etiquetado, un análisis cualitativo de los datos muestra que la presencia de ciertas EN no se corresponden con la patología que hemos de predecir. Por ejemplo, en el caso de sospechas, dudas o negaciones, el informe puede contener una EN como ``Se sospecha una \textit{queratosis} aparente que resulta ser un bulto maligno''.
\end{enumerate}

Buscamos resolver este problema, para lo cual presentamos un método híbrido que combina los transformadores con un modelo basado en RoBERTa \cite{carrino-etal-2022-pretrained} y ontologías \cite{schriml2022human}. Con este fin, creamos con ellos modelos en cascada: detectan el tipo (síntoma, proceso neoplásico, etc.), el sitio anatómico y la gravedad de la patología y un último modelo que, gracias a los anteriores, predice qué patología es. Los informes utilizados provienen de EHRs y son notas clínicas de pacientes escritas por médicos, que pueden ser de primera cita o un seguimiento. Cada informe tiene dos etiquetas asociadas: la patología y una codificación de la patología. Estos informes provienen de la unidad de dermatología de distintos centros de salud de España. Han sido anonimizados de manera semi automática con técnicas basadas en reglas simbólicas. 
Las contribuciones que realizamos con el trabajo actual son las siguientes: 
\begin{itemize}
 \item Un conjunto de datos anonimizado de EHR de dermatología en español, público y de libre acceso\footnote{\url{https://huggingface.co/fundacionctic/DermatES}}.
 \item Una nomenclatura de patologías dermatológicas que viene a enriquecer las ontologías y léxicos existentes. 
 \item Un método híbrido basado en transformadores con ontología para la tarea de clasificación de las EHR con respecto a las patologías posibles.
\end{itemize}

Además de la introducción en la Sección \ref{intro}, el artículo se divide en otras cuatro secciones. En la Sección \ref{statof} se presenta un estado de la cuestión donde resumimos tanto los métodos que se asemejan al nuestro como los recursos lingüísticos que existen. En la Sección \ref{method} se proporciona una descripción de nuestra metodología y de la arquitectura final del modelo. La Sección \ref{xps} se centra en 
los resultados. La Sección \ref{concl} aborda la discusión, conclusiones, y los posibles trabajos futuros.

\section{Estado de la cuestión} \label{statof}

La minería de texto en informes clínicos es un campo importante del PLN desde hace años \cite{mccray1987role}. Sin embargo, la cantidad de trabajos relacionados con la extracción de información en el ámbito médico ha florecido con la expansión de los EHR \cite{pmid24870142}. 

En este sentido, a principios de los 2000 se solía utilizar una combinación de ontología y de web semántica para extraer nombres de enfermedades o de medicamentos \cite{lambrix2005towards,Jing2022.05.11.22274984}. Además, las ontologías se utilizaban combinadas con algoritmos estadísticos \cite{Shannon2021-he}. 

La mayoría de los trabajos en este campo son en inglés, pero algunos como \cite{PLN2721,roma2009ontofis} emplearon textos en español. A partir de 2010 se empezaron a usar redes neuronales como los LSTM \cite{hochreiter1997long}, dado que tienen capacidad para retener relaciones entre las frases. 
Estas redes se utilizaron para el NER médico tanto en lengua inglesa \cite{luo2017recurrent} como española \cite{giner2022reconocimiento}. 

Aun así, los trabajos que buscan asociar la totalidad del texto a un concepto de patología son escasos en comparación con los que aspiran a detectar las patologías nombradas.

En conjuntos de datos en inglés encontramos MIMIC \cite{Johnson2023} y MIMIC-III \cite{Johnson2023-fn}. Por otro lado, en español está el conjunto privado \cite{10.1007/978-3-031-34953-9_38}, SPACCC (\textit{Spanish Clinical Cases Corpus}) que no contiene etiquetado \cite{intxaurrondo2019spaccc} o PharmacoNER \cite{gonzalez-agirre-etal-2019-pharmaconer} pero también para NER. El conjunto en español que más se asemeja al del presente trabajo es el CodiEsp \cite{miranda2020overview} pero está etiquetado en diagnóstico y procedimiento. 
Para consultar conjuntos de datos existentes, se puede consultar la amplia lista creada por \cite{10.1093/jamia/ocy173}. 

En general, siendo los datos de salud sensibles, y con la necesidad de cumplir el RGPD\footnote{\url{https://www.hacienda.gob.es/es-ES/El\%20Ministerio/Paginas/DPD/Normativa\_PD.aspx}} \cite{9420457}, es imprescindible anonimizar los datos clínicos \cite{iglesias2008mostas}. Como métodos explorados de anonimización, \cite{Lordick2022} usan BRAT \cite{stenetorp-etal-2012-brat} para anonimizarlos. \cite{lima-lopez-etal-2020-hitzalmed} muestra que un modelo híbrido hace casi imposible la reidentificación de las personas. En conjunto con estos trabajos, nos inspiramos en \cite{francopoulo:hal-02939437} y MEDOCCAN \cite{marimon2019automatic} para anonimizar los nuestros. 

Por otro lado, respecto a los métodos que clasifican el texto entero en vez de realizar NER, encontramos a \cite{krishnan2019ontology}, que emplea ontologías para predecir las enfermedades en textos clínicos. \cite{shoham2023cpllm} aplica los transformadores inspirados en BERT \cite{devlin-etal-2019-bert} para crear encajes léxicos médicos. Los trabajos que ofrecen mejores resultados en términos de exactitud y precisión son los que combinan la potencia lingüística de los modelos de lengua masivos (LLM) y de las ontologías, por lo que nos inspiramos en estos para diseñar nuestro trabajo. Por ejemplo, \cite{9679070} implementaron transformadores BERT y ontologías médicas, obteniendo los mejores resultados sobre MIMIC. Hasta donde alcanza nuestro conocimiento, no existe un trabajo en español que describa cómo predecir la patología de un paciente a partir del EHR textual utilizando estos métodos.

\section{Metodología} \label{method}

En esta sección presentamos el conjunto de datos que creamos y procesamos, así como la técnica de anonimización que usamos para proteger la privacidad de su contenido. Finalmente, describimos los modelos que utilizamos además de nuestro método híbrido: transformador-ontología con modelos en cascada.

\subsection{Descripción del conjunto de datos}

\begin{figure*}
 \centering
 \includegraphics[width=\linewidth]{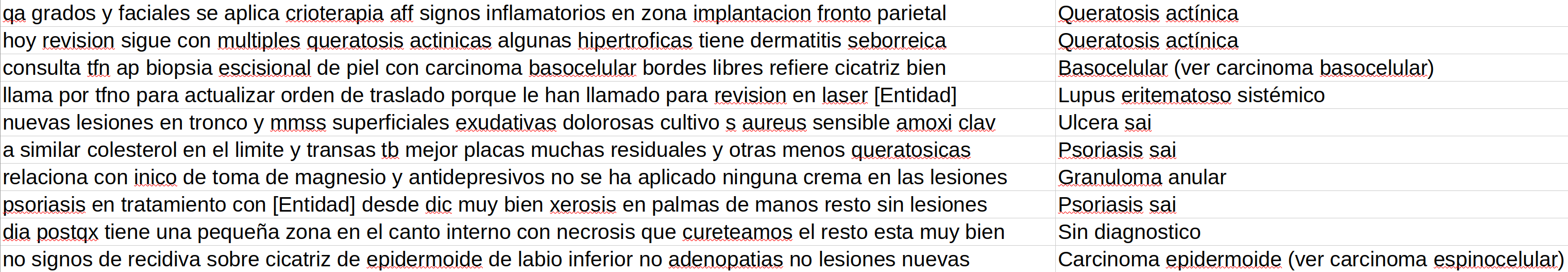}
 \caption{Ejemplo del conjunto de datos. A la izquierda,  el informe de primera consulta o de seguimiento. A la derecha, la patología a predecir.}
 \label{fig_data}
\end{figure*}

Los datos utilizados se corresponden con notas clínicas de hospitales españoles con respecto a consultas de dermatología, tanto de primera consulta como de revisiones posteriores. Dichos datos vienen dados en ficheros individuales en formato HL7 (\textit{Health Level 7}) \cite{Saripalle2019-pz}, el cual se trata de un conjunto de estándares internacionales que permiten el intercambio, integración, compartición y recuperación de datos electrónicos de salud. Además, facilita que la comunicación entre diferentes sistemas sea más ágil y fiable. 
Nuestro corpus contiene 8881 informes y 173 patologías dermatológicas diferentes. Cada informe contiene una sola etiqueta de una patología dermatológica y estamos ante un caso de clasificación multiclase. Los informes clínicos ligados a 43 variables de diversa índole, incluidas el nombre y el código de la patología. Dado el objetivo planteado en este proyecto, se ha limitado dicho conjunto a dos variables de interés: el texto escrito por parte del facultativo sobre la consulta en lenguaje natural, y la variable que ofrece la patología diagnosticada al paciente dentro de la taxonomía considerada por el sistema de recogida. Un ejemplo del conjunto de datos está ilustrado en la Figura \ref{fig_data}. 

Teniendo en cuenta el número de patologías y su reparto desequilibrado (Figura \ref{distr_enf} en Anexo \ref{an_ic}), intuíamos que incluso con nuestro método híbrido, muchas clases iban a ser obviadas durante el entrenamiento del modelo. Es por eso que estudiamos definir el umbral del mínimo de ejemplos por patología que maximice la precisión del modelo sin perder demasiadas patologías. Decidimos guardar las 25 patologías más representadas que corresponden a un mínimo de 61 ejemplos por categoría. En el anexo \ref{an_umbral} explicamos que ese umbral es el óptimo para conservar cierto número de categorías sin mermar la eficacia de los modelos y en el anexo. 

El conjunto de datos no ha sido lematizado o pasado por otros preprocesamientos clásicos tal y como tampoco lo hicieron \cite{tornberg2023use} explicando que los modelos de lenguaje hoy en día son capaces de procesar textos brutos.

\subsection{Anonimización}

A lo largo de este apartado se detalla el proceso de anonimizado del conjunto de datos. Los datos médicos pueden presentar información sensible que no ha de compartirse. Por esta razón, se ha procedido con diferentes fases semi-automatizadas que permitan enmascarar información de carácter privado. 
En un primer nivel, se ha procedido a eliminar todo aquel contenido numérico que aparezca en dichos mensajes. Esto se debe a que dicha información está ligada a información sensible: fechas, años, edades o diferentes identificadores (DNI, identificador del paciente, etc.). 
Por otro lado, se ha abordado la detección de otros tipos de información sensible, que en lugar de eliminar como se ha hecho con los caracteres numéricos, se ha procedido con su enmascaramiento con la etiqueta ``[Entidad]''. En este caso, se han tratado entidades como nombres propios, apellidos, ciudades/localizaciones o nombres de hospitales.
Para llevar a cabo este paso, se ha procedido con la identificación de diferentes fuentes externas que permitan localizar tal información, haciendo uso de las siguientes:
\begin{itemize}
 \item Lista de apellidos y nombres de hombre y mujer más frecuentes en España, proporcionados por el INE (Instituto Nacional de Estadística)\footnote{\url{https://www.ine.es/dyngs/INEbase/es/operacion.htm?c=Estadistica_C&cid=1254736177009&menu=resultados&idp=1254734710990}}.
 \item Lista de palabras más frecuentes en el lenguaje español mediante recursos proporcionados por la Real Academia Española (RAE) ligadas al Corpus de Referencia del Español Actual (CREA)\footnote{\url{https://corpus.rae.es/lfrecuencias.html}}.
 \item Lista con las ciudades y hospitales más habituales dentro de la fuente de datos utilizada.
\end{itemize}

Así, el proceso de enmascaramiento se ha desarrollado del siguiente modo:
\begin{itemize}
 \item Se han identificado todas aquellas apariciones de los nombres propios de hombre o mujer más frecuentes.
 \item Se han identificado todas aquellas apariciones de los apellidos más frecuentes como primer apellido.
 \item Entre dichos nombres y apellidos, se filtran aquellos que estén entre los términos más frecuentes para que no sean filtrados.
 \item Se añaden un total de 43 excepciones que puedan ser relevantes para el ámbito particular de la dermatología (\textit{cabello}, \textit{seco}, \textit{benigno}, etc.). 
 \item Se enmascaran patrones que han sido identificados como candidatos a contener información sensible (texto posterior a los términos \textit{dr}, \textit{dra}, \textit{doctor}, \textit{doctora}).
\end{itemize}

Con esto, se genera un conjunto de textos anonimizado y enmascarado, que se ha procedido a analizar para validar que la información está protegida. Para ello, se ha realizado una revisión manual por parte de dos revisores:
\begin{itemize}
 \item Se ha seleccionado una muestra del conjunto de datos anonimizado del 10\% del total. 
 \item Se ha realizado dicha selección de forma estratificada con respecto a la categorización del texto original. 
 \item Dicho conjunto se ha dividido a su vez en dos subconjuntos del mismo tamaño, donde cada uno de ellos tiene entradas únicas y entradas compartidas entre sí.
\end{itemize}

\begin{figure}
 \centering
 \includegraphics[width=\linewidth]{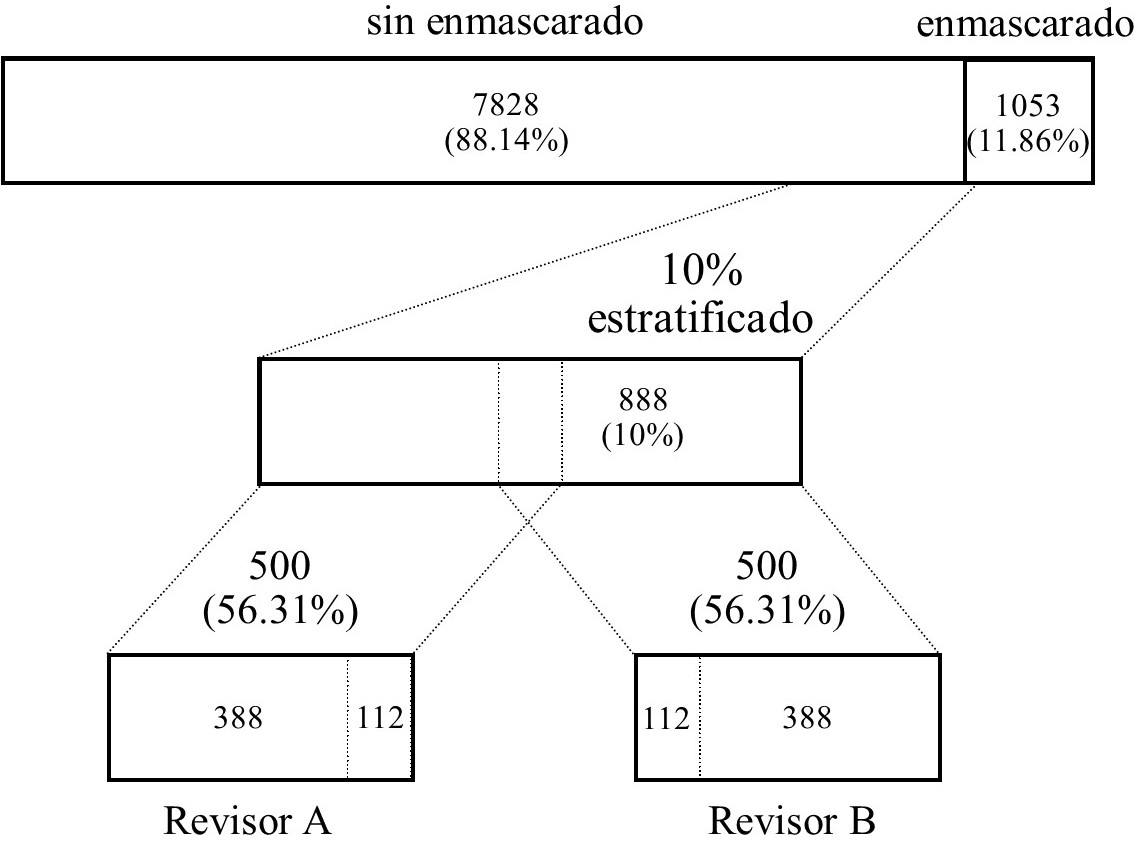}
 \caption{Representación gráfica de la partición generada para la validación de la anonimización realizada.}
 \label{particion}
\end{figure}

En la Figura \ref{particion} se presenta dicha partición generada, así como los tamaños exactos de cada uno de los conjuntos.


Así, se ha validado si el anonimizado de cada texto ha sido adecuado o si se ha equivocado protegiendo información innecesaria o no protegiendo información sensible.

Tras dicha revisión, se han identificado casos particulares a corregir y patrones generalizados. Se han corregido con iteraciones posteriores análogas para garantizar la correcta protección de dicha información. 

A nivel de acuerdo interjueces, de las 112 observaciones comunes, los revisores solo han discrepado en 4. Dichos errores han sido analizados y subsanados en el proceso semi-automático previo para todo el conjunto. 

\subsection{Modelo de lenguaje}

Dada la potencia y la cantidad de datos necesarios para entrenar un modelo de lenguaje \cite{kim2024breakthrough}, la opción más eficiente para poder usar transformadores es realizar \textit{fine-tuning}. El objetivo de esto es encontrar el modelo preentrenado que mejor se ajuste a nuestros datos y a la detección de una patología. Para este problema, elegimos el bsc-bio-ehr-es\footnote{\url{https://huggingface.co/PlanTL-GOB-ES/bsc-bio-ehr-es}}, un modelo preentrenado de lenguaje biomédico-clínico diseñado para el idioma español. El modelo ha sido preentrenado utilizando datos de textos biomédicos y clínicos en español para aprender patrones lingüísticos específicos. Se basa en la arquitectura de RoBERTa \cite{liu2019roberta}. 
Sin embargo, los resultados (presentados en la Sección \ref{subs_res}) evidenciaron la dificultad que tiene un único modelo para aprender a detectar enfermedades, por lo que enriquecimos el entrenamiento usando información de ontologías médicas y varios modelos en cascada, donde cada uno aprende informaciones específicas de lo datos.

\subsection{Ontologías utilizadas y modelos en cascada}

A lo largo de esta sección, abordamos en un primer nivel el tratamiento del desequilibrio de la variable objetivo, seguido por la revisión de las ontologías médicas y traducción aplicadas, con un apartado final centrado en los modelos en cascada propuestos.

\subsubsection{Tratamiento del desequilibrio de clases}


El desequilibrio presente en los datos implica que la mayoría de las clases tienen una cobertura casi nula y provocan un sobreentrenamiento del modelo con las tres primeras clases. 
Para remediar este problema intentamos reducir la dimensionalidad no de manera matemática con PCA \cite{wadud2022can} o T-SNE \cite{liu2021identifying}, sino con modelos en cascada, cada uno tratando de resolver una tarea más simple y agregando su salida a la entrada del siguiente modelo, hasta ser capaz de predecir la patología exacta. 
Nos inspiramos en \cite{dinarelli-rosset-2011-models} y \cite{10.1007/978-3-030-87802-3_19}, quienes utilizaron este método para el NER  y para el reconocimiento de voz, respectivamente.  
Por otro lado, en vez de usar métodos probabilísticos para reducir la variabilidad de las clases, introducimos determinismo en la arquitectura de nuestro método mediante ontologías que permiten extraer el tipo de patología, el sitio anatómico afectado, la gravedad o la intensidad. Esto nos permite reagrupar las patologías en relaciones semánticas mas genéricas que consiguen mejorar la precisión a la hora de predecir la patología en cada informe. 

\subsubsection{Ontologías médicas y traducción}
 Hubieron trabajos previos que combinaban aprendizaje automático y ontologías \cite{ghidalia2024combining}, pero dadas las características tanto lingüísticas como de dificultad de la tarea, nuestro método es original en términos de extracción de información y de combinación de modelos especializados.
Aunque existen ontologías en lengua española como ONTERMET \cite{vila2015ontoloxias} o ECIEMAPS \cite{villaplana2023improving}, tienen el inconveniente de ser demasiado especializadas o poco completas. Por este motivo decidimos traducir de forma automática \cite{stahlberg2020neural} el nombre de las patologías de nuestro conjunto de etiquetas del español al inglés con la API Google Translate y usar ontologías médicas más generales y completas como UMLS \cite{bodenreider2004unified}, SNOMED \cite{spackman1997snomed}, MedDRA \cite{brown1999medical} y HumanDO \cite{schriml2019human,schriml2022human}.



Tras esto, se ha accedido a dicha información a través de las bibliotecas de Python \textit{PyMedTermino} y \textit{PyMedTermino2} \cite{Lamy2015-pu}, así como \textit{medcat} \cite{Kraljevic2021-ln} , diseñadas para acceder a estas ontologías. Con estas herramientas, identificamos la codificación correspondiente a cada ontología de las enfermedades analizadas de forma semisupervisada, revisando que la identificación se ajuste a la enfermedad real y no a posibles variaciones similares. 

Analizando las características y metadatos de dichas ontologías, extrajimos varios metadatos relevantes. Primero, utilizando SNOMED es posible identificar diferentes sitios anatómicos de la patología a través del \textit{finding site}. Luego, usando de manera combinada UMLS, ICD-10 y MedDRA, extrajimos el tipo y la gravedad de la enfermedad.
Para extraer estas características nos hemos inspirado en la clasificación de características predictivas dermatológicas propuesta en \cite{Fisher2016-fe}. 

\subsubsection{Modelos en cascada}

Los modelos en cascada aprenden las relaciones mencionadas en la Tabla \ref{type_transl} (ampliada en la Tabla \ref{type_transl_an} del Anexo \ref{an_ic}).

\begin{figure*}
 \centering
 \includegraphics[width=0.8\linewidth]{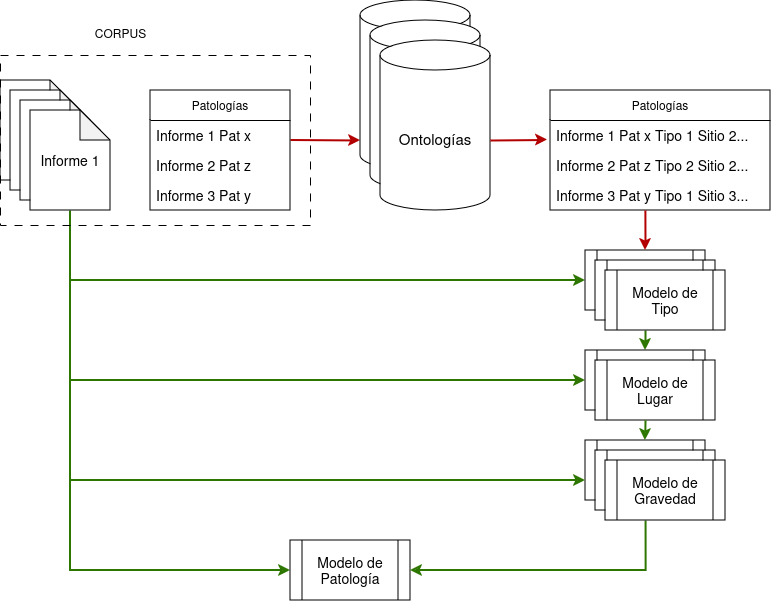}
 \caption{Arquitectura de nuestro método (en rojo las etapas solo de entrenamiento, en verde las de entrenamiento e inferencia).}
 \label{fig_satec}
\end{figure*}

Una vez que hemos extraído de los conceptos (las etiquetas traducidas) las relaciones comunes existentes en la ontología, entrenamos el bsc-bio-ehr-es para predecir cada una de ellas (ver Tabla \ref{type_transl}). Cada relación predicha sirve para predecir la siguiente. En la Sección 4 mostramos en qué orden deben de ser aprendidas las relaciones. La Figura \ref{fig_satec} ilustra nuestro método: 
 
\begin{enumerate}
 \item Extraemos del corpus el nombre de las patologías, las traducimos, las convertimos en conceptos y recuperamos las relaciones ligadas a esos conceptos dentro de las ontologías.
 \item Por cada relación extraída, entrenamos un modelo. En modo entrenamiento, las relaciones son un oráculo. En modo inferencia, las relaciones son generadas por el modelo.
 \item Cada modelo intermediario recibe como entrada los informes y una relación del modelo previo. 
 \item En cada etapa de la cascada, es decir cuando un modelo ha realizado su predicción, se descodifica cada una de ellas y se concatena con el informe inicial, y se vuelve a vectorizar el conjunto informe-relación predicha mediante el tokenizador del bsc-bio-ehr-es. En la ultima etapa, el vector de entrada contiene una representación del informe y de las tres relaciones.
 \item Un modelo final aprende a predecir la patología a partir de los informes y de la salida del último modelo intermediario. 
 
\end{enumerate}


\begin{table*}
\caption{\label{type_transl} Nomenclatura generada a partir de las características extraídas con \textit{Pymedtermino} (muestra de las 5 enfermedades más frecuentes, versión completa con todas las enfermedades y frecuencias de aparición en la Tabla \ref{type_transl_an} del Anexo \ref{an_ic})}
\begin{center}
\begin{tabular}{|l|l|l|l|}
\hline
Enfermedad & Tipo & Gravedad & Sitio\\
\hline
carcinoma de células basales & proceso neoplasico & importante & piel \\ 
  psoriasis & proceso autoinmune & inofensivo & extremidades \\ 
  nevus melanocítico & precancer & inofensivo & todo \\ 
  acné & enfermedad & leve & todo \\ 
  queratosis actínica & precancer & inofensivo & piel \\ 
\hline
\end{tabular}
\end{center}

\end{table*}

\section{Experimentos y resultados} \label{xps}

En esta sección describimos los resultados de nuestra arquitectura comparada con modelos de referencia, y evaluamos su rendimiento.


\subsection{Enriquecimiento con ontologías}

Esta estrategia consiste en entrenar un modelo previo al de clasificación de informes: se trata de un modelo intermediario para aprender el tipo de enfermedad. Para eso, usamos UMLS \cite{bodenreider2004unified} para recuperar los tipos de patología, SNOMED para el sitio anatómico y ICD10 para la gravedad\footnote{Se descartó MedDRA por no aportar más que las otras}. Traducimos de manera automática el nombre de las enfermedades con Google Translate. 

\subsection{Modelos}

Se han considerado diferentes enfoques para este modelado, tanto haciendo uso de técnicas basadas en transformadores, como modelos de aprendizaje supervisado de clasificación. 

En cuanto a aquellos basados en transformadores, se han utilizado los siguientes:
\begin{itemize}
 \item BETO \cite{CaneteCFP2020}. Tal y como indican sus autores, se trata de un modelo BERT entrenado con un corpus en español, utilizando la técnica \textit{Whole Word Masking}. 
 \item \textit{bsc-bio-ehr-es} \cite{carrino-etal-2022-pretrained}. Modelo generado por el BSC (\textit{Barcelona Supercomuting Center}), utilizando de base un gran corpus de textos biomédicos en español.
\end{itemize}

Además, el segundo de estos modelos ha sido tratado a través del tuneado de hiperparámetros. Se especifican esos hiperparámetros y el material usado para entrenar los modelos en el Anexo \ref{hyper_and_stuff}.

En lo que respecta a modelos clásicos de aprendizaje automático de clasificación, se ha recurrido a tres de los algoritmos más habituales: la regresión logística, las máquinas de vector soporte (SVM) y los \textit{Random Forest}. 




Se han propuesto dos enfoques: 
\begin{itemize}
    \item  El modo \textbf{oráculo} (OR): el modelo\footnote{\url{https://huggingface.co/fundacionctic/oracle-dermat}} conoce el tipo, el sitio anatómico y la gravedad de la patología. El objetivo de este modo es demostrar la necesidad de información externa para realizar la tarea de clasificación.
    \item  El modo \textbf{predictivo} (PR): el modelo final\footnote{\url{https://huggingface.co/fundacionctic/predict-dermat}} debe predecir las tres características mencionadas, en el orden óptimo. Cada inferencia de una característica debe ayudar a la predicción de la siguiente. El objetivo de este modo es demostrar que nuestro modelo tiene una aplicación real y útil para la comunidad médica. 
\end{itemize}

\subsection{Resultados y evaluación}\label{subs_res}

De cara a generar una comparativa de OR y PR, hemos seleccionado las siguientes métricas: 

\begin{itemize}
    \item Exactitud. Proporción de predicciones correctas sobre el total de predicciones realizadas. 
    \item F1-score. Media armónica de la precisión y el \textit{recall}. Para problemas multiclase, el F1-score puede calcularse de manera macro (media de f-score de cada clase) o micro (media entre falsos positivos, falsos negativos y reales positivos de todo el conjunto).
    \item Exactitud \textit{top-k}. Proporción de casos en los cuales la clase verdadera está entre las k predicciones más probables del modelo. Esta métrica es útil cuando no solo importa la predicción más probable, sino también otras alternativas que el modelo considere razonables. Para todo este artículo, consideramos (k=2).
    
    \item F1-score \textit{top-k}. Extensión del concepto de F1-score considerando las k clases más probables predichas por el modelo.
    
\end{itemize}

Inspirándonos en los trabajos de \cite{NIPS2015_0336dcba,DBLP:journals/corr/abs-2007-15359} en aprendizaje automático y \cite{DBLP:journals/corr/cmp-lg-9706014} en la tarea de POS-\textit{tagging}, consideramos que profesionales médicos usando nuestro modelo pueden ver más informativo tener no una sino dos predicciones (viendo la cantidad de clases posibles), al parecerse esto más al estilo de decisión natural humano, y dejando que sea el médico el que tenga el veredicto final de la enfermedad. 


Estas métricas han sido obtenidas para todas las configuraciones contempladas en esta investigación. Las flechas corresponden a los diferentes modelos en cascada: cada característica predicha se convierte en característica conocida para el siguiente modelo.

\subsubsection{Resultados modelos intermediarios}

Exponemos en la Tabla \ref{interm-table} un resumen de los resultados obtenidos con el bsc-bio-ehr-es en PR sobre cada uno de los tres componentes del sistema de cascada: tipo, gravedad y sitio de la enfermedad. Los detalles de todos los resultados se pueden encontrar en el Anexo \ref{res_inter}. 

\begin{table}
\caption{\label{interm-table} Tabla con los resultados intermediarios para tipo (\textbf{t}), gravedad (\textbf{gr}) y sitio (\textbf{sit}) de cada enfermedad}
\begin{center}
\begin{tabular}{|l|l|l|l|}
\hline
Cat. info. & Prec. &  Micro F1 & Macro F1 \\
\hline
\textbf{t} & 0.57 & 0.56  & 0.38        \\
\textbf{gr} & 0.57 & 0.56  & 0.41     \\
\textbf{sit} & 0.68 & 0.67  & 0.59     \\
\textbf{t} $\rightarrow$ \textbf{gr} $\rightarrow$ \textbf{sit}  & 0.70 & 0.68  & 0.58     \\
\textbf{gr} $\rightarrow$ \textbf{t} $\rightarrow$ \textbf{sit}  & 0.62 & 0.61  & 0.51     \\
\textbf{sit} $\rightarrow$ \textbf{gr} $\rightarrow$ \textbf{t}  & \textbf{0.72} & \textbf{0.71 } &\textbf{0.62}     \\
\hline
\end{tabular}
\end{center}

\end{table}
La mejor combinación de categorías intermediarias en cascada parece ser primero predecir el sitio de la enfermedad, seguido por la gravedad y por fin el tipo. Observemos ahora si se refleja en la predicción final de la enfermedad. 

\subsubsection{Modelo final de predicción de enfermedad}

En todas las combinaciones de experimentos, bien sea con modelos de aprendizaje automático clásicos o con transformadores ajustados, se trata de una tarea de clasificación supervisada multiclase monoetiqueta. 

\begin{table*}
\caption{\label{tabla_resultados} Tabla con las métricas obtenidas para cada configuración considerada (AAC: aprendizaje automático clásico; TR: transformador; PR modo predictivo; OR modo oráculo; bsc bsc-bio-ehr-es)}
\begin{center}
\resizebox{\textwidth}{!}{
\begin{tabular}{|l|l|l|l|l|l|}
\hline
Modelo & Precisión & Micro F1-sc. & Macro F1-sc & Prec. \textit{top-k} & F1-sc. \textit{top-k}\\
\hline
Regresión logística (AAC) & 0.25 & 0.16 & 0.12 & 0.37 & 0.31\\
SVM (AAC) & 0.263 &  0.13  & 0.14 & 0.39 & 0.29\\
Random Forest (AAC) & 0.268 & 0.19 & 0.12 & 0.40 & 0.33\\
\hline
PR BETO (TR) & 0.34 & 0.38 & 0.12 & 0.63 & 0.60\\
PR bsc (TR)  & 0.52 & 0.50 & 0.42 & 0.67 & 0.61\\
PR bsc (TR) + \textbf{gr} $\rightarrow$ \textbf{sit} $\rightarrow$ \textbf{t} & 0.58 & 0.55 & 0.47 & 0.69 & 0.62\\
PR bsc (TR) + \textbf{sit} $\rightarrow$ \textbf{gr} $\rightarrow$ \textbf{t} & 0.61 & 0.59 & 0.53 & 0.67 & 0.59\\
PR bsc (TR) + \textbf{t} $\rightarrow$ \textbf{gr} $\rightarrow$ \textbf{sit} & 0.63 & 0.60 & \textbf{0.54} & 0.68 & 0.61\\
PR bsc (TR) + \textbf{t} $\rightarrow$ \textbf{sit} $\rightarrow$ \textbf{gr} & \textbf{0.66} & \textbf{0.61} & 0.38 & \textbf{0.71} & \textbf{0.65}\\
\hline
OR bsc (TR) + \textbf{t} & 0.64 & 0.47 & 0.42 & 0.78 & 0.63\\
OR bsc (TR) + \textbf{gr} & 0.55 & 0.53 & 0.36 & 0.69 & 0.60\\
OR bsc (TR) + \textbf{sit} & 0.65 & 0.63 & 0.50 & 0.75 & 0.74\\
OR bsc (TR) + \textbf{sit} $\rightarrow$ \textbf{t} & 0.77 & 0.76 & 0.66 & 0.87 & 0.87\\
OR bsc (TR) + \textbf{t} $\rightarrow$ \textbf{sit} $\rightarrow$ \textbf{gr} & \textbf{0.84} & \textbf{0.82} & \textbf{0.75} & \textbf{0.92} & \textbf{0.90}\\
\hline
\end{tabular}
}
\end{center}

\end{table*}

Observando la Tabla \ref{tabla_resultados}, se puede apreciar cómo los mejores resultados con respecto a las cuatro métricas se obtienen para el modelo basado en ontologías con todas las informaciones añadidas, existiendo una diferencia sustancial con el resto de opciones. En OR, cuando el modelo conoce las informaciones de las ontologías, los resultados superan el 0.84 de precisión absoluta y el 0.92 en precisión \textit{top-k}. En PR, la ganancia también es significativa puesto que pasamos del 0.5 de precisión con el modelo ``vanilla'' a 0.66 con la mejor combinación de información. Es notable que la mejor combinación de características en cascada sea la de tipo seguido por sitio y gravedad, puesto que la gravedad solo tiene 4 variables, lo que nos hacía intuir que su aprendizaje sería más sencillo. Estos resultados confirman nuestras intuiciones: la necesidad de buscar informaciones externas para el entrenamiento del modelo y la necesidad de buscar en qué orden hay que aprender estas informaciones para optimizar el sistema final de clasificación, así como la eficacia del bsc-bio-ehr-es para nuestra tarea. En la Tabla \ref{disease-table} se presentan los mejores resultados en OR y PR para las 5 patologías más frecuentes. 

\subsection{Análisis de errores de los modelos}

\begin{table}
\caption{\label{disease-table} Tabla con las métricas (sin \textit{top-k}) para las 5 patologías mas frecuentes}
\begin{center}
\begin{tabular}{|l|l|l|}
\hline
Enfermedad & F1 PR & F1 OR \\
\hline
acné             & 0.86 & 0.94        \\
carc. cél. basales     & 0.70 & 0.92        \\
psoriasis           & 0.81 & 0.87        \\
nevus melanocítico       & 0.72 & 0.93        \\
queratosis actínica       & 0.63 & 0.83        \\

\hline
\end{tabular}
\end{center}

\end{table}

Tras la generación de estos modelos se ha procedido a realizar un análisis de errores de la mejor configuración PR y del mejor OR. 

Por un lado, la poca precisión de los tres modelos clásicos de aprendizaje supervisado (regresión logística, SVM, \textit{Random Forest}) tiene como posible causa la incapacidad de generalizar sobre etiquetas poco frecuentes.
La principal limitación de los modelos clásicos de \textit{machine learning} está en su falta de memoria de contexto: cuanto más largo sea un texto, más costoso es para la máquina recordar el contenido del principio. Las altas dimensiones de \textit{embeddings} tampoco ayudan, puesto que la cantidad de variables a aprender es exponencial.

Examinando las categorías de enfermedad donde más erran los modelos, concluimos que las discrepancias son debidas a que las enfermedades confundidas comparten zonas del cuerpo afectadas similares, niveles de gravedad parecidos, y algunas descripciones de aspecto visual y síntomas compartidos. Por otro lado, a excepción de los cánceres, el modelo tiende a confundir enfermedades cuya apariencia física son las protuberancias.
La confusión más frecuente es entre el carcinoma de células basales y de células escamosas, representando 844 de 2334 errores. Aunque pueden aparecer en cualquier parte del cuerpo, es más probable que estas enfermedades se desarrollen en áreas expuestas al sol, como la cabeza, el cuello y los brazos. La diferencia clave entre ellos es su gravedad, siendo el carcinoma escamoso más agresivo.
Otra confusión frecuente del modelo es el acné con la queratosis seborreica con 325 errores. 
Estas enfermedades tienen similitudes en aspecto visual (protuberancias, enrojecimiento, picazón) y en lugares de aparición (cara y torso). Son dos condiciones dermatológicas que pueden parecerse en textos descriptivos, lo que podría llevar a confusión en un modelo de lenguaje.


\section{Discusión, conclusión y líneas de trabajo futuras} \label{concl}

De manera general, los errores cometidos por el modelo pueden ser explicados, pese a la dificultad de la tarea (muchas enfermedades posibles, informes médicos que pueden ser tanto de primera cita o de seguimiento). 

Debido a estos errores hemos decidido usar una métrica basada en el \textit{top-k} en vez de la exactitud estricta. Si este sistema se utiliza en un entorno médico, entendemos que un profesional prefiere tener que elegir entre 2 posibles diagnósticos en vez de 25. 
Es también interesante mencionar que el mejor escenario en PR es cuando el modelo tiene que aprender tanto el sitio anatómico como el tipo y la gravedad de la enfermedad antes de adivinar esta última. Concluimos que un transformador necesita esa información externa para realizar la clasificación porque su modelo de lenguaje no es suficiente. En un primer momento, intuíamos que el modelo que aprende primero de la gravedad daría el mejor resultado, puesto que esta variable solo tiene 4 categorías (inofensiva, leve, moderada, extrema). Sin embargo, el mejor resultado lo da un modelo que aprende primero el tipo de enfermedad, seguido por el lugar y al final la gravedad de la enfermedad. Eso significa, y es lógico a nivel médico, que primero se detecta el tipo de patología, antes de intentar adivinar su gravedad. 

Como conclusiones, nuestro trabajo presenta varias contribuciones importantes. Primero, proporcionamos un nuevo conjunto de datos de EHR en español de dermatología, anonimizado y etiquetado en patologías. Segundo, proponemos un método novedoso para predecir en EHR la patología que padece una persona. 
Tercero, este método es un sistema híbrido compuesto por varios modelos transformadores especializados puestos en cascada que usan como entrada además de los EHR, la salida del modelo precedente. El último modelo es el que predice la patología exacta. 
Los resultados muestran que los modelos en cascada son más eficaces que un solo modelo para distinguir enfermedades poco frecuentes. Eso significa que un solo modelo de extremo a extremo de transformadores no es suficiente para distinguir un concepto de patología entre varias decenas en un EHR, pero además que el uso de ontologías externas es necesario para que los transformadores aprendan conceptos intermediarios relacionados con la patología, de forma similar al aprendizaje humano. Aun así, los resultados muestran un margen de mejora importante. Esto puede abordarse desde diferentes puntos de vista, como el etiquetado manual de los EHR en primera cita o cita de seguimiento, la búsqueda de nuevas relaciones ontológicas (como por ejemplo características de etiología más exhaustivas), o probar modelos diferentes para aprender cada característica. 

Como posible línea de trabajo futuro proponemos automatizar el uso de ontologías mediante el RAG \cite{gao2024retrievalaugmented}, como en BiomedRAG \cite{li2024biomedrag} así como el uso de NegEx \cite{Arguello-Gonzalez2023} para evitar los falsos positivos, acompañado de nuestro sistema de modelos en cascada para eliminar el determinismo a la hora de encontrar relaciones de concepto para convertirse en un modelo de extremo a extremo. 


\begin{acknowledgments}
Agradecimiento a la entidad SATEC por proporcionar el conjunto de datos, a la Fundación CTIC que puso a nuestra disposición todo el material y los recursos. Agradecemos a los revisores así como a Bea de Otto por la lectura y los comentarios pertinentes que mejoraron el articulo. Agradecemos y felicitamos al BSC por el modelo \textit{bsc-bio-ehr-es}.
\end{acknowledgments}

\renewcommand{\refname}{Referencias}
\bibliography{EjemploARTsepln}

\appendix

\section{Detalles experimentales}

En este apartado, explicamos y describimos en detalles los resultados obtenidos y el procedimiento para reproducir el experimento.

\subsection{Algoritmos de extracción de información de las ontologías}
En el Algoritmo \ref{alg1}, explicamos el proceso de extracción de información de las ontologías para agregar relaciones entre cada informe y la enfermedad asociada a éste. 
\begin{enumerate}
    \item Por cada enfermedad, la traducimos en inglés mediante la API de Google Translate.
    \item Luego, por cada ontología en nuestra lista tenemos diferentes relaciones, por lo que es importante usarlas todas. A partir del nombre de la enfermedad traducida en inglés tenemos tres opciones:
    \begin{enumerate}
    
        \item Si es SNOMED, extraemos el sitio del cuerpo y de la piel afectados.
        \item Si es UMLS, extraemos el  por la enfermedad.
        \item Si es ICD10, extraemos la gravedad de la enfermedad mediante la interpretación de la información sobre si es una afección mayor o menor, y si conlleva morbilidad.
    \end{enumerate}
\end{enumerate}

\begin{algorithm}
	\caption{Extracción de relaciones} 
    \label{alg1}
	\begin{algorithmic}[1]
        \State load \textit{pymedtermino} library
		\For {$diseases=1,2,\ldots,M$}
            \State translate {$diseases$} with Google API
			\For {$ontology=1,2,\ldots,N$}
                \If{$SNOMED$}
                    \State {$diseases.getType()$}
                \ElsIf{$UMLS$}
                    \State{$diseases.getLocation()$}
                \Else
				    \State $diseases.getAffection()$
                    \If{$has(minor)$}
                        \State{$SetTo(light)$}
                    \ElsIf{$has(major)$}
                        \State{$SetTo(important)$}
                    \ElsIf{$has(morbidity)$}
                        \State{$SetTo(deadly)$}
                    \Else
                        \State{$SetTo(inoffensive)$}
                    \EndIf
                \EndIf
			\EndFor
		\EndFor
	\end{algorithmic} 
\end{algorithm}

\subsection{Algoritmo de los modelos en cascada}

En este apartado, explicamos el funcionamiento de los modelos en cascada (Algoritmo \ref{alg2}): 
\begin{enumerate}
    \item Realizamos la combinación ordenada de las tres relaciones que tenemos, con paquetes de tamaño 1 o 3.
    \item Por cada uno de esos paquetes, cargamos el dataset y lo tokenizamos.
    \item  Por cada elemento de cada paquete, entrenamos un modelo en modo supervisado para aprender ese elemento.
    \item Una vez que cada modelo está entrenado, su predicción del elemento se concatena con la entrada original, y sirve de nueva entrada para entrenar el modelo a predecir el próximo elemento del paquete.
    \item Entrenamos otro modelo con el próximo elemento del paquete y la nueva entrada.
    \item Cuando todos los elementos han sido aprendido por los diferentes modelos en cascada, se entrena un último modelo con la entrada original enriquecida por la última salida (con todos los elementos predichos) para predecir la enfermedad.
    \item Una vez que todos los paquetes están procesados, se comparan los modelos y se elige el mejor.
\end{enumerate}

\begin{algorithm}
	\caption{Modelos en cascada}
 \label{alg2}
	\begin{algorithmic}[1]
        \State $combi \gets []$
		\For {$iter1=1,2,\ldots N$}
            \For{$iter2=1,2,\ldots M$}
		          \State $combi \gets iter1,iter2$
            \EndFor
        \EndFor
	    \For{$relations$ in $combi$}
            \State $input \gets string(medicalRecords)$
            \State $input.tokenize()$
            \For{$relation$ in $relations$}
                \State $output \gets model.train(relation)$
                \State {$input$ $\gets$ {$input + output$}}
            \EndFor
            \State $output \gets$ $Model.train(diseases)$
        \EndFor
        \State $computeAccuracy()$
        \State $getBestModel()$
	\end{algorithmic} 
\end{algorithm} 

\subsection{Configuración del entrenamiento} \label{hyper_and_stuff}
Se han identificado los siguientes parámetros del modelo como susceptibles de tratamiento:
\begin{itemize}
 \item Tamaño del lote (\textit{batch size}). Número de muestras de entrenamiento que se procesarán a través de la red en una sola iteración antes de que se actualicen los pesos del modelo.
 \item Tasa de aprendizaje (\textit{learning rate}). Controla cuánto se ajustan los pesos del modelo en respuesta al error calculado en cada iteración del entrenamiento.
 \item Número de épocas (\textit{epochs}). Número de veces que el algoritmo de aprendizaje trabajará a través de todo el conjunto de datos de entrenamiento.
\end{itemize}

Para cada uno de estos hiperparámetros se han contemplado diferentes valores posibles y, tras un proceso de \textit{grid search}, se ha determinado que los valores que mejores resultados proporcionan son \textit{batch size} 64, \textit{learning rate} 0.001 y \textit{epochs} 10. 

Los experimentos se llevaron a cabo mediante una NVIDIA GeForce RTX 2080 Ti 12GB y una  NVIDIA RTX A6000 50GB. Usamos la bibioteca Pytorch 2.2.1 con CUDA 12.1. Cada entrenamiento duraba entre 3 y 5 minutos dados los pocos datos en el corpus. En total, contando el \textit{grid search}, los experimentos necesitaron algo mas que 96 horas para obtener los resultados presentados en el artículo. 

\subsection{Resultados detallados de los modelos intermediarios} 
\label{res_inter}
En esta sección, presentamos los resultados de varias combinaciones en la Tabla \ref{interm-fulltable} que se llevaron a cabo para predecir la patología de la mejor manera posible. Tratándose de variaciones sin repetición, el numero total de combinaciones posibles es de la forma: 
\[
\sum_{k=1}^{n} V_n^k = \sum_{k=1}^{n} \frac{n!}{(n-k)!}.
\]

Para \( n = 3 \), el número total de combinaciones (variaciones) de tamaños 1 a 3 sin repetición es:

\[
\frac{3!}{(3-1)!} + \frac{3!}{(3-2)!} + \frac{3!}{(3-3)!} = 15.
\]

\begin{table*}
\caption{\label{interm-fulltable} Tabla con los resultados intermediarios para tipo, gravedad y sitio de cada enfermedad}
\begin{center}
\begin{tabular}{|l|l|l|l|}
\hline
Categoría de información & Precisión &  Micro F1 & Macro F1 \\
\hline
\textbf{t} & 0.57 & 0.56  & 0.38        \\
\textbf{gr} & 0.57 & 0.56  & 0.41     \\
\textbf{sit} & 0.68 & 0.67  & 0.59     \\
\hline
\textbf{gr} $\rightarrow$ \textbf{t} $\rightarrow$ \textbf{sit}   & 0.62 & 0.61  & 0.51     \\\textbf{gr} $\rightarrow$ \textbf{sit} $\rightarrow$ \textbf{t}   & 0.66 & 0.66  & 0.58     \\
\textbf{t} $\rightarrow$ \textbf{gr} $\rightarrow$ \textbf{sit}  & 0.70 & 0.68  & 0.58   \\
\textbf{t} $\rightarrow$ \textbf{sit} $\rightarrow$ \textbf{gr}   & 0.70 & 0.68  & 0.57     \\\textbf{sit} $\rightarrow$ \textbf{t} $\rightarrow$ \textbf{gr}   & 0.69 & 0.67  & 0.58     \\
\textbf{sit} $\rightarrow$ \textbf{gr} $\rightarrow$ \textbf{t}  & \textbf{0.72} & \textbf{0.71 } &\textbf{0.62}     \\
\hline
\end{tabular}
\end{center}

\end{table*}

\subsection{Comparativa entre los resultados de los diferentes enfoques}

Se ha llevado a cabo una comparativa del método OR completo con la información de las ontologías (modelo A) y el método basado en el modelo bsc-bio-ehr-es \textit{vanilla} sin ninguna información añadida (modelo B) como referencia de los desarrollos presentados en esta investigación, haciendo especial hincapié en las mejoras que ha proporcionado el utilizar la visión \textit{top-k}, y viendo en qué situaciones dichos cambios ha supuesto una mejora en los resultados. Esta comparativa es presentada en la Tabla \ref{tabla_umbrales}.

Para el modelo A, las métricas de precisión y F1-score generadas sin el uso de \textit{top-k} se corresponden con 0.82 y 0.73, las cuales se han visto mejoradas con el enfoque \textit{top-k} pasando a valores 0.86 y 0.85. Del mismo modo, se ha observado como el modelo B pasa de valores 0.52 y 0.42 para dichas métricas a 0.67 y 0.61 con la inclusión del \textit{top-k}. Si bien en el modelo A se aprecian mejoras evidentes, es en el caso B donde existe una mejora más sustancial.

\begin{table*}
\caption{\label{disease-table_full} Tabla con las métricas (sin \textit{top-k}) para las 25 patologías mas frecuentes}
\begin{center}

\begin{tabular}{|l|l|l|l|}
\hline
Enfermedad & F1 \textit{vanilla} & F1 PR & F1 OR \\
\hline
acné        & 0.43  & 0.86 & 0.94        \\
carc. cél. basales  & 0.60   & 0.70 & 0.92        \\
psoriasis      & 0.67    & 0.81 & 0.87        \\
nevus melanocítico & 0.52 & 0.72 & 0.93        \\
queratosis actínica   & 0.49 & 0.63 & 0.83        \\
carcinoma de células escamosas  & 0.43  & 0.52 & 0.86        \\
eccema           & 0.45  & 0.59 & 0.62        \\
rosácea      & 0.00 & 0.37 & 0.55        \\
lentigo solar   & 0.54 & 0.65 & 0.97        \\
liquen escleroatrófico   & 0.73 & 0.69 & 0.82        \\
fibroma        & 0.57 & 0.50 & 0.87        \\
llaga          & 0.64   & 0.69 & 0.78        \\
melanoma       & 0.43    & 0.66 & 0.86        \\
alopecia areata   & 0.74  & 0.80 & 0.80        \\
dermatitis atópica   & 0.45 & 0.47 & 0.74        \\
carcinoma espinocelular & 0.40 & 0.60 & 0.91 \\
queratosis seborreica & 0.44 & 0.67 & 0.97 \\
sin diagnostico & 0.31 & 0.37 &  0.77 \\
acné juvenil & 0.04 & 0.00 & 0.63 \\
verruga & 0.57 & 0.82 & 0.98 \\
urticaria crónica & 0.22 & 0.71 & 0.87 \\
hemangioma & 0.86 & 0.77 & 0.97 \\
nevus melanocítico atípico & 0.00 & 0.00 & 0.91 \\
dermatofibroma & 0.53 & 0.53 &  0.95 \\
ulcera & 0.00 &0.55 & 0.70 \\
\hline
\end{tabular}

\end{center}

\end{table*}

Ilustrativamente, se presenta tabla \ref{disease-table_full} un análisis de en qué situaciones ambos modelos han mejorado por el uso del enfoque \textit{top-k}. Con respecto al modelo A, los principales errores vienen de predecir ``dermatitis atópica'' donde la realidad se corresponde a ``eccema'' (17.65\%) o ``psoriasis'' (9.8\%). En el caso de ``eccema'', tiene sentido que se dé dicha confusión, dado que son dos términos que se usan a menudo indistintamente para referirse a la misma afección cutánea. La ``dermatitis atópica'' es el término más específico, mientras que el ``eccema'' es un término más general que abarca varios tipos de inflamación de la piel. En lo que respecta a la confusión con ``psoriasis'', pueden darse razones para ello, como la apariencia similar en ciertas situaciones (enrojecimiento, inflamación y picazón común, pudiendo llegar a escamación en ambos casos), así como en cuanto a la localización corporal (ambas pueden aparecer en áreas como los codos, rodillas o cuero cabelludo). 

En el caso del modelo B, la principal mejora se presenta en los casos en los que la enfermedad es ``nevus melanocítico adquirido'' y se predice ``nevus melanocítico'' (11.89\%). Los nevus melanocíticos adquiridos y congénitos comparten varias características, incluyendo su apariencia, histología, genética y desarrollo. Sin embargo, se diferencian en el momento de su aparición, siendo los congénitos presentes al nacer o en las primeras semanas de vida, mientras que los adquiridos aparecen a lo largo de la vida. Esto ilustra cómo no solo el enfoque \textit{top-k} mejora la precisión de los modelos, sino que algunos de los errores más comunes son entendibles dado el significado de los términos confundidos. 

Cabe destacar también que el enfoque \textit{top-k} permite en varios casos mejorar la predicción que incluye situaciones ``sin diagnóstico'', proporcionando dicha opción en casos donde el enfoque base no lo contempla. También es reseñable cómo en ambos métodos aparece un alto número de casos de confusión entre diagnósticos ligados a carcinoma (``carcinoma de células basales'', ``carcinoma de células escamosas'') y queratosis (``queratosis actínica'', ``queratosis seborreica''). En este caso no son patologías semejantes, por lo que el incluir \textit{top-k} proporcionaría al facultativo una alternativa sobre la que valorar cuál es la opción adecuada con su conocimiento experto. 

\subsection{Resultados del modelo con diferentes umbrales de frecuencia de cada patología} \label{an_umbral}
La Tabla \ref{tabla_umbrales} muestra los resultados tanto del método A como del B, y demuestra que el umbral de 61 ejemplos mínimo por categoría es el óptimo para guardar un máximo de categorías sin perder la eficacia de los modelos de clasificación.

\begin{table*}
\caption{\label{tabla_umbrales} Métricas obtenidas con diferentes umbrales de ejemplos por patología A es el método con la información de las ontologías y B el método con el bsc-bio-ehr-es \textit{vanilla}}
\begin{center}
\resizebox{\textwidth}{!}{
\begin{tabular}{|l|l|l|l|l|l|l|l|}
\hline
Modelo & Umbral & Num. clases & Prec. & Micro F1-sc. & Macro F1-sc & Prec. \textit{top-k} & F1-sc. \textit{top-k}\\
\hline

B & 2 & 173 & 0.39 & 0.34 & 0.08 & 0.54 & 0.13\\
B & 10 & 76 & 0.41 & 0.41 & 0.12 & 0.58 & 0.27\\
B & 25 & 44 & 0.46 & 0.43 & 0.24 & 0.59 & 0.46\\
B & 50 & 27 & 0.48 & 0.46 & 0.38 & 0.66 & 0.63 \\
\textbf{B (nuestro)} & 61 & 25 & 0.52 & 0.50 & 0.42 & 0.67 & 0.61\\
B & 75 & 20 & 0.51 & 0.49 & 0.44 & 0.69 & 0.71\\ 
B & 100 & 15 & 0.55 & 0.54 & 0.51 & 0.71 & 0.77 \\
\hline\hline
A & 2 & 173 & 0.68 & 0.62 & 0.14 & 0.80 & 0.28\\
A & 10 & 76 & 0.72 & 0.66 & 0.25 & 0.86 & 0.52\\
A & 25 & 44 & 0.77 & 0.73 & 0.48 & 0.90 & 0.81\\
A & 50 & 27 & 0.83 & 0.80 & 0.72 & 0.91 & 0.86\\
\textbf{A (nuestro)} & 61 & 25 & 0.84 & 0.82 & 0.75 & 0.92 & 0.90\\
A & 75 & 20 & \textbf{0.87} & 0.85 & 0.80 & 0.94 & 0.91\\
A & 100 & 15 & \textbf{0.87} & \textbf{0.87} & \textbf{0.85} & \textbf{0.96} & \textbf{0.92}\\

\hline
\end{tabular}}
\end{center}

\end{table*}

\section{Información complementaria} \label{an_ic}

\begin{figure*}
\caption{Distribución de las enfermedades en el conjunto de datos generado.}
 \centering
 \includegraphics[width=\linewidth]{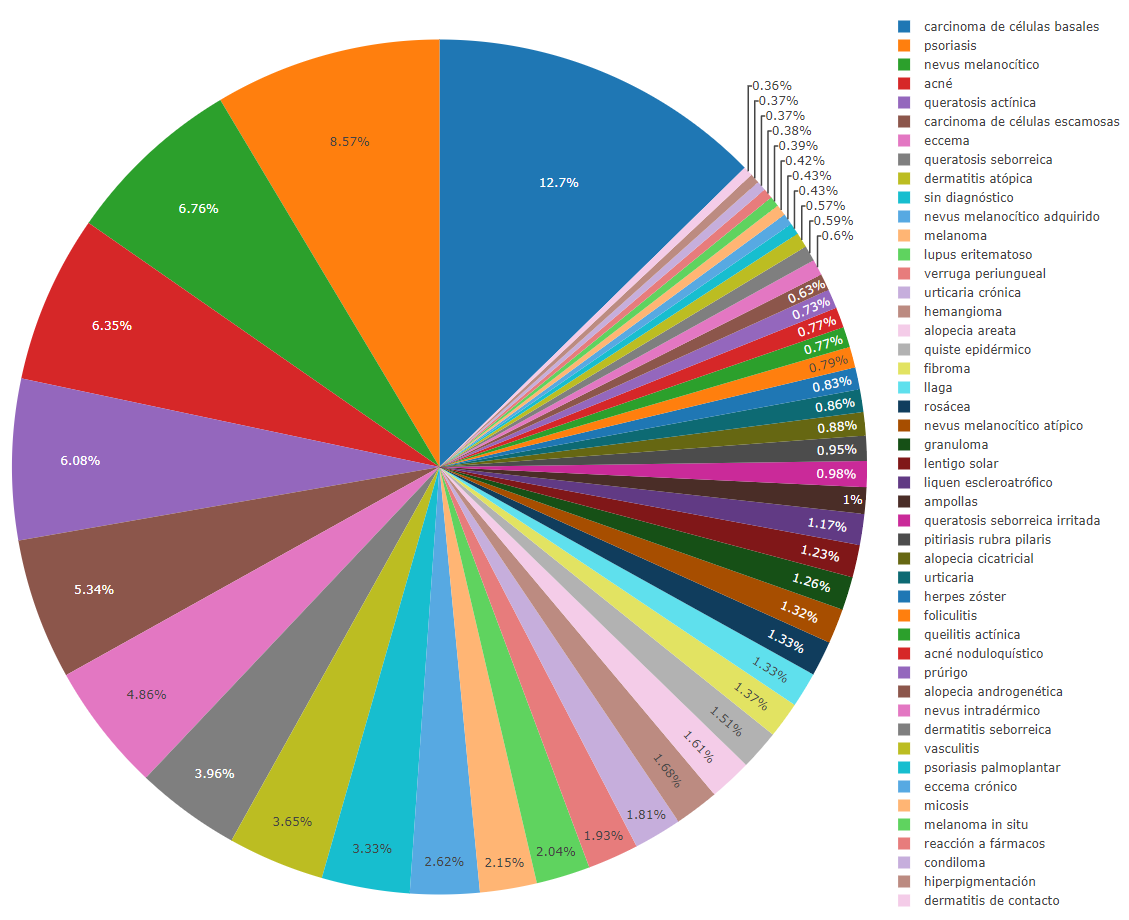}
 
 \label{distr_enf}
\end{figure*}

\begin{figure*}
 \centering
 \includegraphics[width=\linewidth]{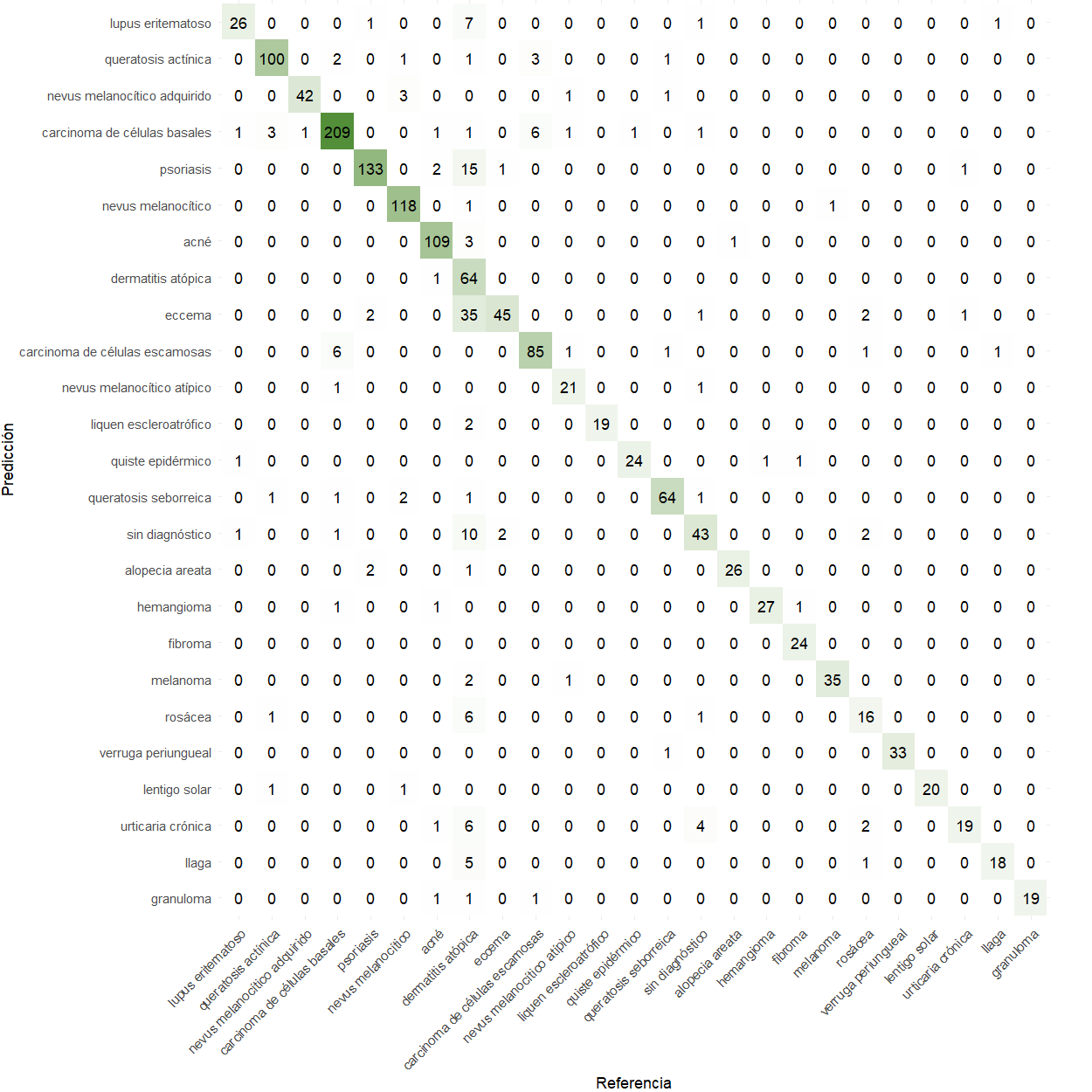}
 \caption{Matriz de confusión para modelo A.}
 \label{cm_full}
\end{figure*}

\begin{figure*}
 \centering
 \includegraphics[width=\linewidth]{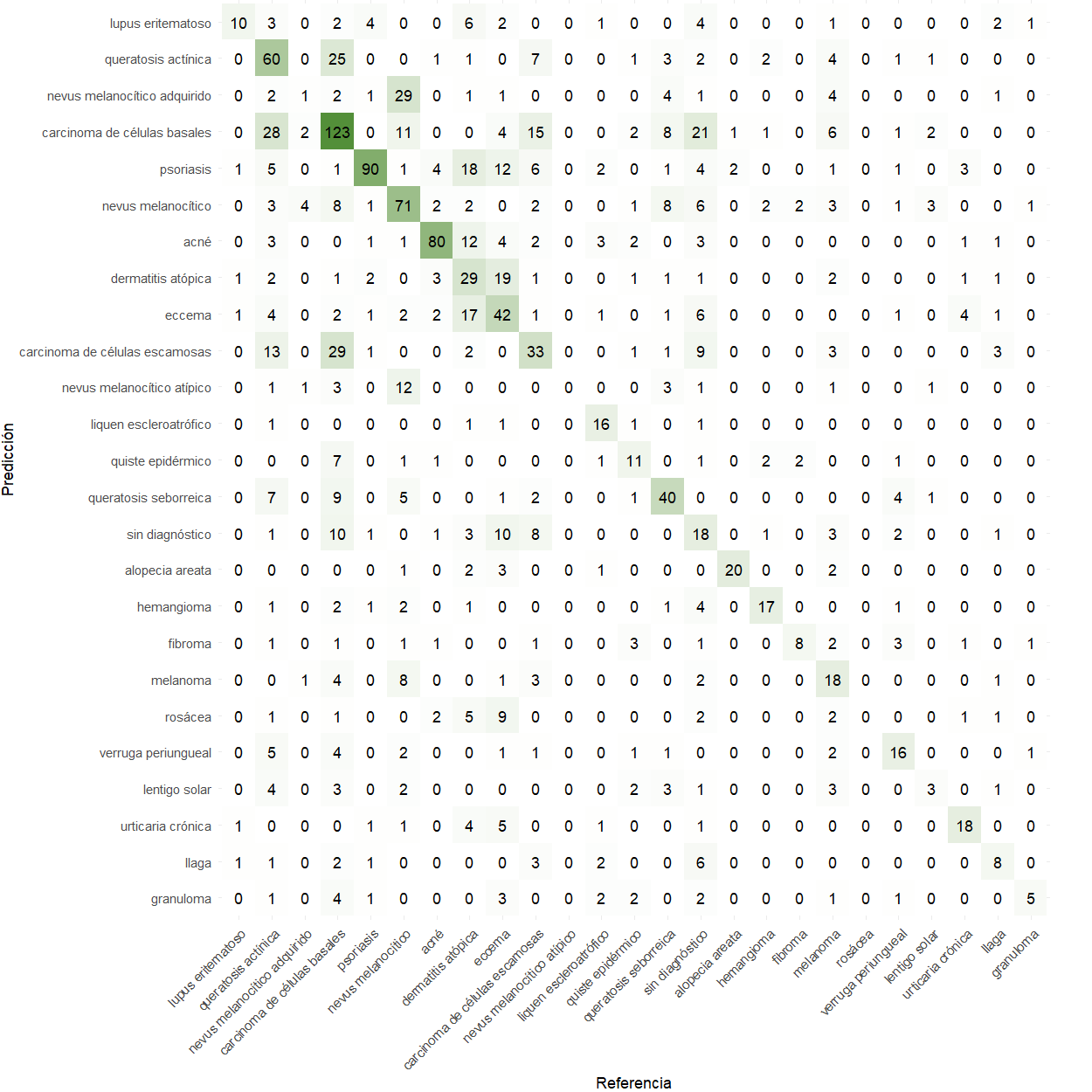}
 \caption{Matriz de confusión para modelo B.}
 \label{cm_vanilla}
\end{figure*}

En esta sección se presenta información complementaria que sirve de apoyo a la comprensión de los desarrollos de este trabajo. En particular, se presenta la siguiente información:

\begin{itemize}
    \item Tabla \ref{type_transl_an}. Versión ampliada de la Tabla \ref{type_transl}, con la lista completa de enfermedades.
    \item Figura \ref{distr_enf}. Representación gráfica de la distribución de las enfermedades en el conjunto de datos.
    \item Figuras \ref{cm_full} y \ref{cm_vanilla}. Matrices de confusión de los métodos A y B presentados en el Anexo \ref{an_ic}, respectivamente. 
\end{itemize}

\begin{table*}
\caption{\label{type_transl_an} Nomenclatura generada a partir de las características extraídas con \textit{Pymedtermino} con las enfermedades mas frecuentes y la cantidad de aparición}
\begin{center}
\begin{tabular}{|l|l|l|l|l|}
\hline
Enfermedad & Tipo & Gravedad & Sitio & Frecuencia\\
\hline
carcinoma de células basales & proceso neoplasico & importante & piel & 1124 \\ 
  psoriasis & proceso autoinmune & inofensivo & extremidades & 761 \\ 
  nevus melanocítico & precancer & inofensivo & todo & 600 \\ 
  acné & enfermedad & leve & todo & 564 \\ 
  queratosis actínica & precancer & inofensivo & piel & 540 \\ 
  carcinoma de células escamosas & proceso neoplasico & extrema & piel & 474 \\ 
  eccema & enfermedad & inofensivo & mano & 432 \\ 
  queratosis seborreica & tumor benigno & inofensivo & piel & 352 \\ 
  dermatitis atópica & enfermedad & inofensivo & articulaciones & 324 \\ 
  sin diagnóstico & sin enfermedad & inofensivo & todo & 296 \\ 
  nevus melanocítico adquirido & proceso neoplasico & inofensivo & extremidades & 233 \\ 
  melanoma & proceso neoplasico & extrema & todo & 191 \\ 
  lupus eritematoso & proceso autoinmune & extrema & tejido conectivo & 181 \\ 
  verruga periungueal & infeccion & inofensivo & mano & 171 \\ 
  urticaria crónica & sintoma & inofensivo & todo & 161 \\ 
  hemangioma & tumor benigno & leve & todo & 149 \\ 
  alopecia areata & proceso autoinmune & inofensivo & cabeza & 143 \\ 
  quiste epidérmico & anormalidad & leve & cara & 134 \\ 
  fibroma & tumor benigno & leve & pierna & 122 \\ 
  llaga & sintoma & inofensivo & boca & 118 \\ 
  rosácea & enfermedad & inofensivo & cara & 118 \\ 
  nevus melanocítico atípico & proceso neoplasico & importante & torso & 117 \\ 
  granuloma & infeccion & extrema & genitales & 112 \\ 
  lentigo pielar & sindrome & leve & todo & 109 \\ 
  liquen escleroatrófico & proceso autoinmune & leve & genitales & 104 \\ 
  ampollas & sintoma & inofensivo & mano &  89 \\ 
  queratosis seborreica irritada & enfermedad & inofensivo & todo &  87 \\ 
  pitiriasis rubra pilaris & proceso autoinmune & leve & articulaciones &  84 \\ 
  alopecia cicatricial & enfermedad & inofensivo & cabeza &  78 \\ 
  urticaria & funcion patologica & leve & todo &  76 \\ 
  herpes zóster & infeccion & importante & torso &  74 \\ 
  foliculitis & enfermedad & inofensivo & cabeza &  70 \\ 
  queilitis actínica & precancer & leve & boca &  68 \\ 
  acné noduloquístico & infeccion & leve & cara &  68 \\ 
  prúrigo & sintoma & inofensivo & cabeza &  65 \\ 
  alopecia androgenética & enfermedad & inofensivo & cabeza &  56 \\ 
  nevus intradérmico & precancer & inofensivo & piel &  53 \\ 
  dermatitis seborreica & proceso autoinmune & inofensivo & cara &  52 \\ 
  vasculitis & proceso autoinmune & extrema & articulaciones &  51 \\ 
  psoriasis palmoplantar & enfermedad & leve & extremidades &  38 \\ 
  eccema crónico & enfermedad & inofensivo & mano &  38 \\ 
  micosis & infeccion & importante & todo &  37 \\ 
  melanoma in situ & proceso neoplasico & inofensivo & todo &  35 \\ 
  reacción a fármacos & envenenamiento & inofensivo & todo &  34 \\ 
  condiloma & infeccion & leve & genitales &  33 \\ 
  hiperpigmentación & anormalidad & inofensivo & todo &  33 \\ 
  dermatitis de contacto & enfermedad & inofensivo & mano &  32 \\ 
\hline
\end{tabular}
\end{center}

\end{table*}

\end{document}